\documentclass[preprint,12pt]{elsarticle}

\usepackage{graphicx}
\usepackage{amssymb}
\usepackage{subcaption}
\usepackage{lineno}

\usepackage{algorithm}
\usepackage{algorithmic}

\journal{Robotics and Autonomous Systems, Elsevier}

\begin{document}

\begin{frontmatter}


\title{Consensus-based Fast and Energy-Efficient \\Multi-Robot Task Allocation}



\author[label1]{Prabhat Mahato}
\author[label1]{Sudipta Saha}
\author[label2]{Chayan Sarkar}
\author[label1]{Md Shaghil}

\address[label1]{School of Electrical Sciences, Indian Institute of Technology Bhubaneswar, India}
\address[label2]{Robotics and Autonomous Systems, TCS Research, India}

\begin{abstract}

In a multi-robot system, the appropriate allocation of the tasks to the individual robots is a very significant component. The availability of a centralized infrastructure can guarantee an optimal allocation of the tasks. However, in many important scenarios such as search and rescue, exploration, disaster-management, war-field, etc., on-the-fly allocation of the dynamic tasks to the robots in a decentralized fashion is the only possible option. Efficient communication among the robots plays a crucial role in any such decentralized setting. Existing works on distributed \textit{Multi-Robot Task Allocation} (MRTA) either assume that the network is available or a naive communication paradigm is used. On the contrary, in most of these scenarios, the network infrastructure is either unstable or unavailable and ad-hoc networking is the only resort. Recent developments in synchronous-transmission (ST) based wireless communication protocols are shown to be more efficient than the traditional asynchronous transmission-based protocols in ad hoc networks such as \textit{Wireless Sensor Network} (WSN)/ \textit{Internet of Things} (IoT) applications. The current work is the first effort that utilizes ST for MRTA. Specifically, we propose an algorithm that efficiently adapts ST-based many-to-many interaction and minimizes the information exchange to reach a consensus for task allocation. We showcase the efficacy of the proposed algorithm through an extensive simulation-based study of its latency and energy-efficiency under different settings.

\end{abstract}


\end{frontmatter}


\section{Introduction}
\label{sec:introduction}

Multi-robot systems have been successfully used in many applications where appropriate allocation of tasks to the fleet of robots plays the anchor role towards the overall efficiency of the systems. Task allocation can be done in both a centralized or decentralized way. In many situations, due to lack of infrastructure, higher possibility of single-point-of-failure, communication bottlenecks, etc., centralized task allocation becomes impractical. Especially the applications like war-field~\cite{gautam2012review}, disastrous scenario~\cite{kano2020decentralized}, underwater exploration~\cite{tsiogkas2015distributed} where any infrastructure is not available, centralized allocation of tasks is not feasible.

In a traditional decentralized task allocation strategy, first, a leader is elected, which oversees the overall task allocation~\cite{gerkey2002sold, giordani2010distributed}. Such strategies will also have similar problems as a centralized solution in hostile scenarios. It may ultimately lead to the frequent election of the leaders, wasting time and energy. Decentralized task allocation without the intervention of a dedicated leader has also been considered in various works~\cite{chopra2017distributed, fang2019collaborative}. Predominantly, decentralized task allocation methods follow various bidding-based schemes. Given a set of tasks, a robot can derive a bid value indicating its suitability in carrying out each of the tasks. Optimal distributed allocation of the tasks can be ensured using a simple two-step algorithm. In the first step, the robots would exchange the bid-values for each task with each of the robots. Next, each robot can run the task allocation algorithm locally as all of them possess the same input (bid value). This naive algorithm involves all-to-all data sharing, which is a very costly operation in terms of the amount of communication needed in the underlying network, especially, when the robots depend on limited battery power. Also, it's very unrealistic to assume that all the tasks are known \textit{a priori}. Tasks are usually explored continuously throughout and hence frequent execution of all-to-all data sharing would be very much necessary for the allocation of tasks.

The other possible way for achieving decentralized multi-robot task allocation is by consensus. A task is supposed to be allocated to a robot that has the highest bid value for that task. Thus in a generic sense, the work of task allocation is finding a network-wide maximum value among the bids. However, the problem becomes complicated since there are multiple tasks, and multiple robots may have got the highest bid value for the same task. Thus, the allocation of a certain number of tasks to a certain set of robots cannot be solved trivially by finding the maximum bid in each of the tasks. All-to-all sharing becomes useful in such a scenario as more information can help break the ties.

To realize both all-to-all data sharing as well as consensus, the underlying communication mechanism, and network protocols play a very crucial role. In most of the existing works either it is assumed that there exist efficient communication mechanisms by which all robots can easily talk to each other or the whole issue related to communication has been ignored. In summary, the multi-robot task allocation (MRTA) problems have not been studied so far in a context where both the allocation and communication protocols are optimized. 

Communication protocols under traditional \textit{asynchronous-transmission} (AT) can manage the multi-robot communication up to a certain load. However, in hostile situations like a war-field or disaster management, an MRTA process needs to maintain quite a good connectivity among the robots over multiple hops as well as under severe constraints of time as well as limited energy in the robots. Under such situations, when the robots need to communicate their task bid-values with each other, they need to use a network-wide one-to-all communication through flooding. In general flooding under AT-based protocols wastes a lot of time and energy due to repeated collisions among uncoordinated transmissions of packets from different nodes (robots). The performance of the system drastically degrades when a series of such one-to-all flooding operations are carried out from all the robots.

In the context of \textit{Wireless Sensor Networks} (WSN) and \textit{Internet-of-Things} (IoT) the communication protocols face similar constraints. \textit{Concurrent/ synchronous-transmission} (ST) based communication protocols have shown great success in many recent works in achieving both high reliability and low-latency under low-energy consumption \cite{syncsurvey}. The solutions under ST have been also shown to work several times faster than AT-based strategies. Therefore, in this work, we find ST-based protocols as the best candidate to fulfill the communication need while carrying out decentralized task allocation in a multi-robot setting.

We propose a bidding-based distributed task allocation mechanism that utilizes an ST-based efficient communication framework, called Chaos~\cite{landsiedel2013chaos} for information exchange during task allocation. Chaos can perform both network-wide aggregation of data as well as all-to-all data sharing. It provides a fast and efficient communication alternative than the corresponding AT-based solution. However, the MRTA solution using Chaos-based all-to-all data sharing requires every robot to share the bidding information for every available task. As a result, the total information exchange is in the order of the number of tasks multiplied by the number of robots. In a distributed bidding-based task allocation strategy, the bid values proposed by individual agents are necessary only to calculate the maximum bid value for each task. It is not necessary for each agent to know the individual bid values proposed by each of the agents for each of the tasks. The ST-based protocol used in our work uses local communications among pairs of nodes to compare the bid values with each other and converges to a global maximum value which is disseminated to all the agents in the network. Moreover, our strategy combines multiple instances of such maximum finding strategies together in an efficient way so that the best bid for each of the tasks becomes known to each of the agents quickly. Therefore, our proposed mechanism ensures maximal information exchange using minimum message exchange, which is also sufficient to ensure an efficient task allocation. The main contributions of this work are summarized in the following.
\begin{itemize}
    \item To the best of our knowledge, this is the first effort that utilizes ST based communication primitive for MRTA.
    \item We propose an efficient adaptation of an ST-based protocol called Chaos that ensures faster and energy-efficient consensus, a.k.a. convergence of information sharing for distributed task allocation.
    \item We develop a smart bidding mechanism that achieves minimal makespan in a distributed task allocation setup using minimal information sharing in the network. It also ensures lower energy consumption by all the robots.
\end{itemize}

The rest of the paper is arranged as follows. Section \ref{sec:relatedwork} provides a comprehensive summary of the existing types of task allocation algorithms. Section \ref{sec:design} describes the design of the proposed algorithm as well as the naive approaches to solve the problem. Section \ref{sec:simulation} provides a detailed simulation based evaluation study of the proposed algorithm. Before concluding, in Section \ref{sec:discussion} several important issues regarding ST and MRTA have been discussed for further clarifications.

\section{Related works}
\label{sec:relatedwork}

There exists a large body of prevalent works that deal with multi-robot task allocation (MRTA) in different scenarios. A complete taxonomy of various MRTA scenarios is summarised by Khamis \textit{et al.}~\cite{khamis2015multi}. Though various centralized task allocations algorithms are proposed in the literature for different classes of MRTA problem~\cite{sarkar2018scalable, agarwal2019cannot}, their applicability is limited to a certain environment where stable communication infrastructure is available. In contrast to the centralized approaches of MRTA, decentralized approaches are proposed to overcome most of the limitations of a centralized system. Since there is a limited research focus on distributed task allocation along with an efficient distributed communication protocol design, we briefly review the relevant literature from both domains. 

\subsection{Distributed task allocation}
One of the earliest distributed MRTA systems is proposed by Gerky \textit{et al.}~\cite{gerkey2002sold} called \textbf{MURDOCH}. It is based on the publish/subscribe mechanism for communication, which puts a heavy load on inter-robot communication. It has been designed by changing the contract net protocol. It uses an auction-based method where first, an auctioneer introduces the task into the system. Then, negotiation is done among different robots who are capable of executing that task. A distributed version of the \textit{Hungarian method} have been proposed Chopra \textit{et al.}~\cite{chopra2017distributed}. In this, each robot performs a sub-step of the centralized version of the Hungarian algorithm for linear sum assignment problem (LSAP). They solve the algorithm in multiple iterations and each iteration robot performs a step of the algorithm in a synchronous manner. It produces an optimal solution in $O(R^3)$ iterations. The communication complexities are much high in this method as more information needs to be shared more often. 

Similarly, many researchers have explored classical distributed systems for MRTA~\cite{giordani2010distributed, giordani2013distributed, das2015distributed}. The generic approaches share information with each other and build a consensus among the competing nodes (robots)~\cite{choi2009consensus, zitouni2020distributed}. The consensus-building often requires to be performed dynamically when tasks and environment are not known a priori or partially known~\cite{ayanian2012decentralized, dasgupta2011multi}. Many evolutionary distributed approaches like market-based task allocation, swarm optimization, ant colony optimization, etc., are also explored in the literature~\cite{trigui2014distributed, wang2016task, liu2013optimal, oh2017market}. In general, these distributed approaches reduce the centralization of computation load to the extent possible. But still, they depend on the underlying communication model, which can lead to inefficiency if not handled properly.

\subsection{Communication protocols for multi-robot task allocation}
There exists a few studies that focus on communication aware distributed task allocation~\cite{liu2015communication, woosley2018integrated, best2018planning, turner2017distributed}. However, these works assume a generic network without consideration of any actual communication protocol. For example, Turner \textit{et al.}~\cite{turner2018fast} proposed a fast consensus mechanism for distributed task allocation and assumes the underlying communication would support the required information exchange. 
The power of energy-efficient communication of a wireless sensor network (WSN) is utilized by a few multi-robot setup~\cite{batalin2004using, batalin2005sensor} as well. The network infrastructure or rather the ad-hoc network setup of a WSN is employed to achieve a distributed task allocation~\cite{lee2014ad,shih2014cooperation}. WSNs are further explored for coordination among a swarm of robots~\cite{han2008swarm, ivanov2019distribution} as it provides an easy and efficient communication mechanism in a multi-robot setup. However, the traditional AT-based protocols in WSN are designed to focus on delay-tolerant applications such as data collection. These protocols would face a challenge in MRTA setup especially when there are quite a few tasks that need to allocate as more tasks require more information exchange within the network.

An alternate paradigm of ST-based low power and fast communication protocols is gaining popularity recently. ST mechanism achieves low duty cycle by switching off the radio most of the time; thus higher energy-efficient communication. Ferrari \textit{e al.}~\cite{ferrari2011efficient} demonstrated a network-wide fast time synchronization and flooding protocol using ST. They utilized a phenomenon called constructive interference (CI) to achieve concurrent transmission and achieved a highly efficient communication primitive where it eliminates a separate medium access control layer and routing layer. Rao \textit{et al.}~\cite{rao2016murphy} proved that CI-based communication can be a viable alternative for embedded systems. As a result, beyond the network-wide flooding, CI-based protocols are developed for upstream data collection in wireless sensor network~\cite{ferrari2012low} as well as bidirectional sensing and control traffic in IoT~\cite{sarkar2019fleet}.

For MRTA, information sharing and consensus-building is required. This involves -- sharing information among each other (participating robots), processing it, and then sharing the result back. Landsiedel \textit{et al.}~\cite{landsiedel2013chaos} proposed an ST-based communication framework for effective information sharing, called Chaos. It can not only be used for all-to-all sharing but its main advantage is, it can do on-the-fly aggregation with all the nodes (robots) in one single period. However, Chaos was not designed for MRTA scenarios, which requires separate consensus building for each of the tasks. As a result, it can not be used as it is for task allocation purposes as it may overburden the network. We adopted the basics of Chaos and coupled it with a smart bidding technique and information sharing strategy to achieve a highly efficient distributed multi-robot task allocation framework.

From the existing literature, it is evident that communication-aware task allocation is a necessity in a distributed setup. Though several custom communication protocols are developed for the applications of wireless sensor networks, a.k.a., IoT, they are not suitable for distributed MRTA. Yet, no major focus is given to developing efficient (custom) protocol design for distributed MRTA. This work focuses on bridging the gap, i.e., building an efficient communication framework for MRTA by utilizing the efficiency of ST.

\begin{figure*}
    \centering
    \includegraphics[width=\linewidth]{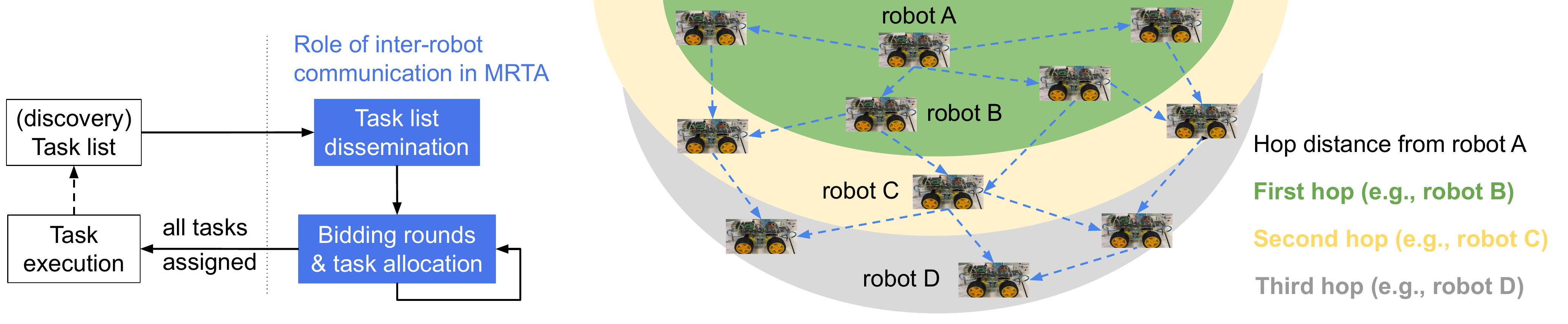}
    \caption{Role of inter-robot communication in distributed multi-robot task allocation (left-side) -- the problem magnifies when no communication infrastructure is available, and inter-robot data sharing is possible only over a multi-hop ad-hoc network (right-side).}
    \label{fig:multi_hop_comm}
\end{figure*}

\section{Design} 
\label{sec:design}

In this section, we describe the design of our distributed task allocation system that utilizes ST-based communication. Figure~\ref{fig:multi_hop_comm} depicts the role of inter-robot communication in distributed task allocation. In particular, the underlying communication protocol plays a role in disseminating the task information about a newly discovered task and placing the bids for task allocation (shown on the left side of Figure~\ref{fig:multi_hop_comm}). Note that all robots may not be able to communicate with each other directly due to the limited transmission range of the radio. For example, the transmission range of agent A is shown using the green region (right side of Figure~\ref{fig:multi_hop_comm}). Thus, even though the wireless communication is broadcast by nature, any transmission by agent A will be received only by its immediate neighbors, e.g., agent B. Immediate neighbors of a node, i.e., nodes that are within communication range are shown using blue dashed arrows. As a result, agent D is reachable to agent A over multiple hops (in this case 3 hops) through other agents and vice-versa. 

In our system, each agent calculates a bid value for every task using a utility function as described in~\cite{sarkar2018semantic}. These bid values are shared using the concurrent transmission-based communication protocol called Chaos~\cite{landsiedel2013chaos}. Although Chaos provides an efficient mechanism for many-to-many interaction in multiple hops, the main challenge lies in efficiently carrying out bidding for multiple tasks in a compact form. In addition, the solution should also support situations when the tasks are being discovered on the go. In the following, we describe three possible strategies for achieving the goals. Different strategies lead to a different amount of information exchange within the network, which defines the energy efficiency and convergence time of task allocation as well as the quality of the allocation. 

\subsection{All-to-All data sharing based Task Allocation (AATA)}
In AATA, every robot shares the bidding information for all the tasks available so far with every other robot in the system using Chaos-based all-to-all data sharing. This is a decentralized version of the centralized strategy where all the information becomes available to all the robots very quickly. However, due to the size limit of a data packet, it requires a lot of time a.k.a., Chaos rounds to share all the bid values. Once all the bid values are exchanged with everyone, a good allocation is done like a centralized algorithm. But, surely the convergence time, as well as the energy consumption, is a concern for this strategy.

\subsection{Independent Bidding Task Allocation (IBTA)}
In this strategy, each robot shares bid value for a single task at a time. We adapted the method proposed by Turner \textit{et al.}~\cite{turner2018fast} for Chaos-based consensus strategy so that a quick agreement can be made among all the robots regarding the maximum bid value. And the task allocation is done using the greedy approach of allocating to the maximum bidder. Unlike the previous strategy where every robot first shares the bid value for every task using all-to-all sharing, IBTA uses consensus when the bid value for a single task is shared by all. This reduces the amount of information exchange and requires fewer rounds to complete the process. However, due to the greedy approach, it leads to non-optimized allocation.

\subsection{Dynamic Bidding Based Task Allocation (DBTA)}
\label{sec:DBTA}
In this strategy, a robot shares more than one bid with all other robots. Sharing of more information brings betterment in the allocation quality and hence a compact overall execution of the tasks is expected. However, an increase in the number of bids also increases the communication among the robots which takes more time for the consensus process to complete. Thus, there is a trade-off between the quality of task allocation and the time and energy consumed for the process itself. The general version of DBTA is referred to as DBTA$_i$, where $i$ indicates how much bid information is shared by each robot. In this paper, by default DBTA refers to DBTA$_2$, which we mainly study and evaluate in comparison to IBTA and AATA. However, we also carry out a study on the effect of various values of $i$ on performance.

\begin{algorithm}
\caption{DBTA-Consensus Algorithm}
\begin{algorithmic}[1]
\STATE $T$: Total tasks
\FOR{all available tasks} 
\STATE Calculate bid value for each task
\STATE Add pair ($\mathit{BidValue, taskId}$) to the $\mathit{Bids}$ list
\ENDFOR
\STATE Sort($\mathit{Bids}$ list based on $\mathit{BidValue}$)
\IF{Initiator}
\FOR{$i \in\mathit{Bids}[0,\; T-1]$}
\IF{$i_{th}$ task is not completed}
    \STATE Place the bid for $i_{th}$ task
    \STATE Break
\ENDIF
\ENDFOR
\ELSE
\FOR{$i \in\mathit{Bids}[0,\; T-1]$ \AND $\mathit{bidPlaced}<2$}
\IF{$i_{th}$ task is not completed} 
    \IF{$i_{th}$ task is in packet}
        \IF{bid is high \AND no previous overbids}
            \STATE Change the bid for $i_{th}$ task
        \ENDIF
    \ELSE
        \STATE Place the bid for $i_{th}$ task at the end of packet
    \ENDIF
\ENDIF
\ENDFOR
\ENDIF
\end{algorithmic}
\end{algorithm} 
The algorithm DBTA is divided into two separate parts -- one for dynamic bidding during consensus (Algorithm 1), and the other for allocating the task using the consensus result (Algorithm 2). In other words, during the bidding and consensus building, how to place the bids is decided using Algorithm 1. And after the convergence, i.e., no new information to share with others, Algorithm 2 is used to decide, who wins the bid for which task. The efficiency of the overall process lies in dynamic bidding where nodes on the fly build the consensus and utilize an intelligent bidding mechanism. A robot avoids bidding for a task when there is no chance of winning the bid, i.e. when it already sees a higher bid for a task than its own bid, it avoids bidding for it altogether even if the task has high bid value for itself. This reduces the amount of information exchange, which in turn helps the process converge faster and saves energy.

\textbf{Algorithm 1} is used for dynamic bidding and consensus of results in the network. In this list, $Bids$ are used to keep the tasks sorted in order of preference for the respective robot such that while placing the bids during consensus the robots can choose their preferences quickly. The algorithm has two sections, one for the initiator and the other for the rest of the robots present in the system, as the process starts with the initiator placing the bids with no partial information. The information used for bidding will be more at the end of the consensus process.

\textbf{Algorithm 2} does the final task allocation based on the consensus results for that particular round. Every robot will first allocate the task with priority 1 to the respective winner of that task. Then the task with priority 2 will be allocated to their respective winner if that robot has not been allocated a task with priority one already. This completes the allocation process.

 \begin{algorithm}[ht!]
\caption{DBTA-Task Allocation Algorithm}
\begin{algorithmic}[1]
\FOR{All the bids present in the packet}
\IF{Winner not allocated in this round \AND $priority$ is 1}
    \STATE The task will be allocated to the winner robot
\ENDIF
\ENDFOR
\FOR{All the bids present in the packet}
\IF{Winner not allocated in this round \AND $priority$ is 2}
    \STATE The task will be allocated to the winner robot
\ENDIF
\ENDFOR
\end{algorithmic}
\end{algorithm}

\subsection{Efficient adaptation of Chaos for distributed task allocation: A compact working of solution}
\label{sec:algo}
As mentioned earlier, Chaos works based on concurrent transmission where multiple nodes are allowed to transmit data at the same time following a very strict microsecond-level time synchronization. When two transmissions from two different nodes carry exactly the same content, they constructively interfere with each other resulting in a correct reception in the receiver. However, when the packet contents differ, the transmission that satisfies the requirements of capture effect \cite{landsiedel2013chaos} wins and the receiver is able to correctly decode the corresponding packet. In addition to this, in Chaos, to help fast dissemination of data, a node refrains from transmitting unless it has got any new information in its previous reception.

\begin{figure*}[t!]
    \centering
    \includegraphics[width=\textwidth]{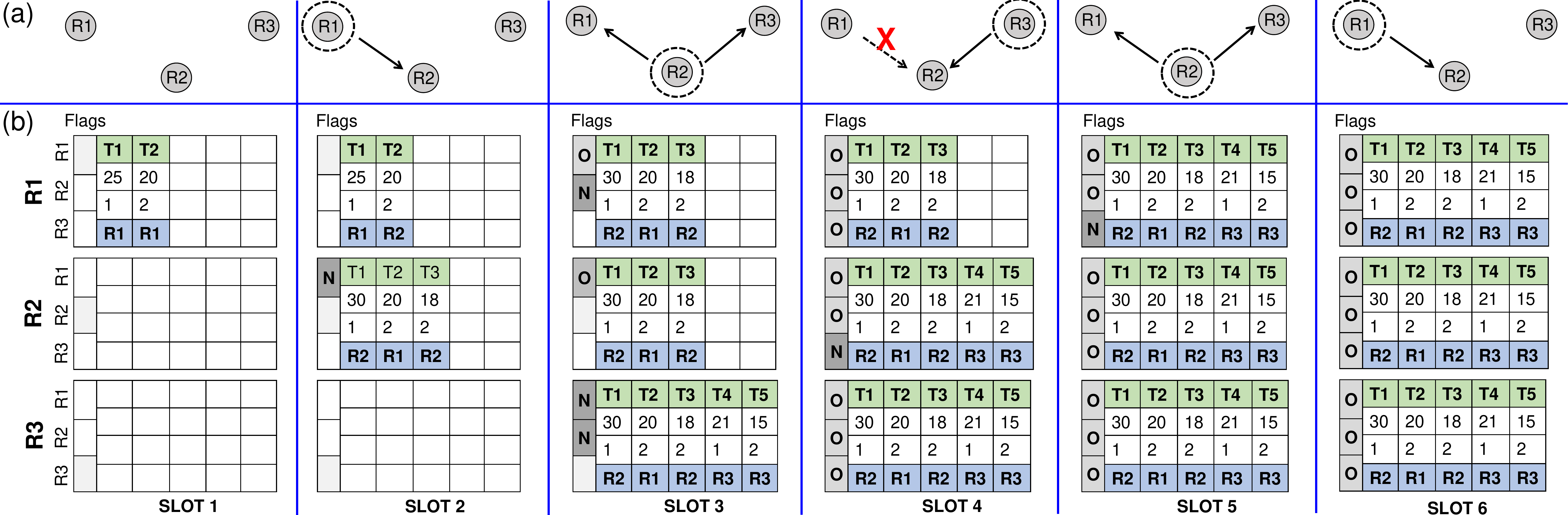}
    \caption{Execution steps of the proposed task allocation strategy through a simplified example. After every data transmission, it is analyzed whether new information is received (marked with N) or not (mark O signifies no new information). A node transmits only if it receives new information.}
    \label{fig:how}
\end{figure*}

Fundamentally, task allocation is done through system-wide agreement on the maximal bid values for each task in a compact form. We use a market-based approach to solve the problem in a fully decentralized fashion. The overall allocation strategy is depicted pictorially in Figure~\ref{fig:how} through an example where five tasks (T1, T2, T3, T4, and T5) are allocated over three robots (R1, R2, and R3). The network topology and the communication between the three robots are depicted in part (a) of the figure. Each robot maintains a data structure for storing the task allocation information, which gradually converges to the final allocation. The same data structure is conveyed through the transmission of the packets. The packet contents in each slot in each robot are shown in Part (b) of the figure in the form of tables. The left-most column of each table is for the flags from each robot to indicate whether a packet received from the robot contains new information (marked by N) or not (marked by O). Followed by this, there are four rows in a table. From top to bottom in sequence, the rows represent the \textit{task ids}, \textit{bid values}, \textit{task priority}, and \textit{id of the robot} to which a task is allocated. 

There are two main parts of the generic packet structure of Chaos. First the set of flags and second the data. For the flags, one bit of information is used for each robot while for the data part it depends on what data are being shared. In our case data is the bid value. When a packet is received by some node, the node is supposed to get the information from the flags that who are nodes already contributed in the aggregation process (in our case maximum value finding). Thus, before transmitting packets based on locally available information a node updates all the flag information. To bootstrap this process, every node sets its own flag in the set of the flags in the packet before transmitting its first packet.

In DBTA, a robot is allowed to bid on two tasks dynamically based on the partial information available during the consensus. Each robot first calculates bids for all the available non-allocated tasks and then selects the best two bids to communicate to others. Priority 1 and 2 are assigned to the best and second-best bids, respectively. All the robots have their own flag set initially. In SLOT 1, let's assume, robot $R1$ initiates the transmission of its packet with the two highest bids. The packet is received by $R2$. In SLOT 2, $R2$ overbids for task $T1$ and places another bid for task $T3$ with priority 2. In SLOT 3, $R2$ transmits and its received by both $R1$ and $R3$. Hence, they both receive new information. $R3$ has no common task available in the packet so it places a bid for two new tasks $T4$ and $T5$. In SLOT 4 both $R1$ and $R3$ transmit their packets but due to the capture effect, only $R3$'s transmission was received by $R2$. $R2$ merges this packet with its own and re-transmits it as it has received new information. In SLOT 5 both $R1$ and $R3$ receive the transmission from $R2$. $R3$ keeps silent as it has no new information to communicate. In SLOT 6 thus, only $R1$ transmits as it has new information to share. At this stage, all the robots come to a consensus as all have no new information to share. Now every robot first allocates the task with priority 1 to the respective winner of that task. Then the task with priority 2 will be allocated to their respective winner if that robot has not been allocated a task with priority 1 already. This completes the process with the allocation of T2 to R1, T1 \& T3 to R2, and T4 \& T5 to R3.
\section{Evaluation}
\label{sec:simulation}

We carry out rigorous experiments with the proposed allocation strategies in the Cooja simulator~\cite{osterlind2006cross}. Under an area of 1000x1000 square meters, we assume that all the robots are initially located near the origin, i.e., location [0, 0], and the tasks are distributed randomly in the entire area. We experimented with a varying number of robots and tasks. Due to the paucity of space, we present only the result sets with 25 robots and up to 140 tasks. Each setup is run for 15 iterations with varying task locations in each iteration and the average values are presented. Before going into the detailed results, let us briefly describe the Cooja simulator, the evaluation metrics and the parameters that we varied across experiments, and the task types.

\begin{figure}[ht!]
    \centering
    \includegraphics[width=\linewidth]{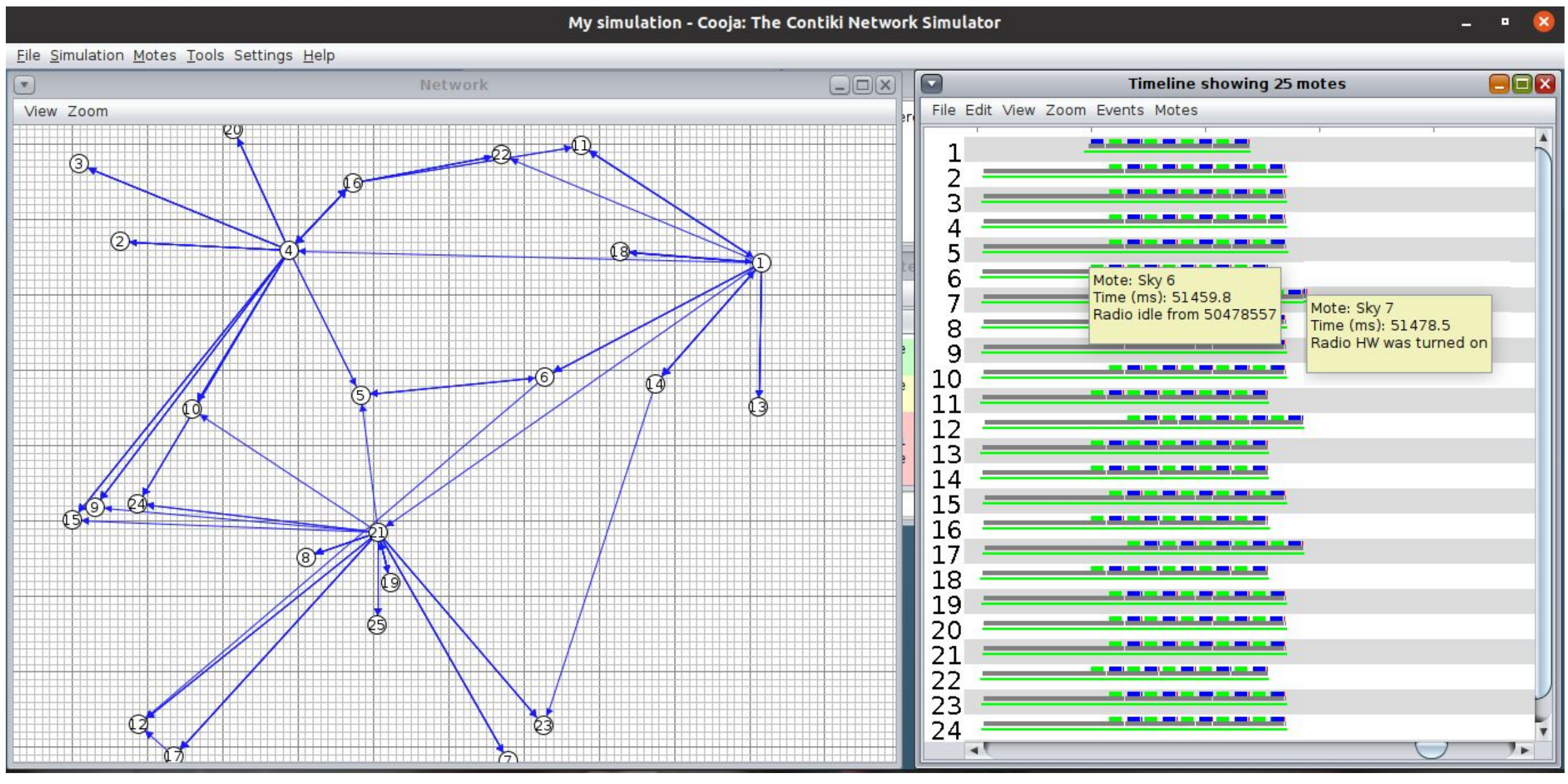}
    \caption{A snapshot from the Cooja simulator with 25 nodes in the simulation. The left window shows data transmission from the nodes (blue arrows signify transmitter-receiver pairs). The right side window shows the timeline of nodes in terms of radio-on, indicated by the grey line, and data transmit and receive, indicated by the small blue and green rectangles, respectively. All the nodes receive the flooding (network-side broadcast) message initiated by node 1 within 20ms (first transmission at 51459.8ms and last one at 51478.5ms) even if they are multiple hops away.}
    \label{fig:cooja1}
    \hspace{2em}
    \includegraphics[width=\linewidth]{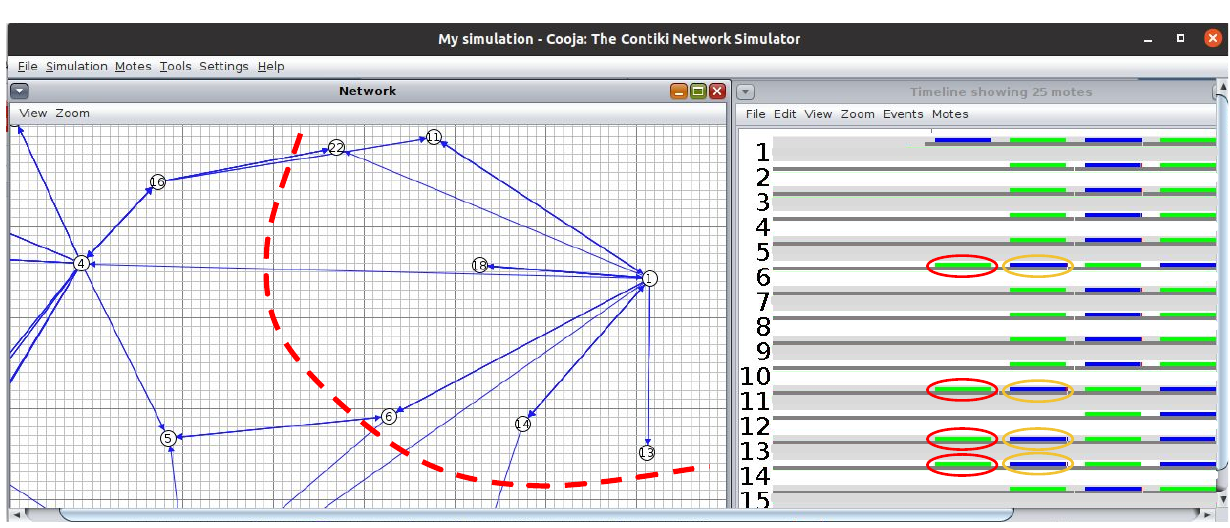}
    \caption{A focused view of Figure~\ref{fig:cooja1} that shows the immediate neighbors of node 1 within the red dotted circular area (left window). When node 1 transmits, only its immediate neighbors receive the data (red circles in the right window) and then they transmit immediately and synchronously (yellow circles in the right window).}
    \label{fig:cooja2}
\end{figure}

\subsection{Cooja: The Contiki Network Simulator}
Cooja is a network simulator for the Contiki operating system~\cite{dunkels2004contiki} that enables cross-level simulation. It enables simultaneous experimentation at multiple levels of the system ranging from low-level simulation of sensor node hardware to higher application-level behavior -- all in the same simulation. It supports multiple micro-controller based platforms, operating system software, radio transceivers, and radio transmission models. Additionally, it is flexible and extensible at all levels of the system. Figure~\ref{fig:cooja1} shows a typical simulation setup wherein the left window shows how messages are exchanged for any set of nodes and any communication protocol. In the right side window, the simulation timeline for all the nodes along with the precise timing of message transmit and receive is shown. 

Figure~\ref{fig:cooja1} depicts a screenshot of one of our experimental setups with 25 nodes (robots). Please note that the number of nodes can be varied easily in a simulation setup (just adding or removing a node instance). Also, it is fully flexible to select among the supported node types and one can test any communication protocol. Figure~\ref{fig:cooja2} zooms on the node 1 and its immediate neighbors from the screenshot as in Figure~\ref{fig:cooja1}. It showcases that when a node transmits a data packet (in this case node 1), it is received only by its immediate neighbors or the nodes that are within its communication range. Since the constructive interference (CI) based communication protocol is employed, the immediate neighbors transmit the received data immediately and synchronously so that the message can reach further away nodes from node 1.

\subsection{Parameters and metrics for evaluation}
We vary the following parameters to evaluate different metrics stated in the latter part of this subsection. 
\begin{itemize}
    \item \textbf{No of robots}: Total number of robots available in the system should be varied.
    \item \textbf{No of tasks}: Total number of tasks should be varied as it is an important factor for the allocation process.
    \item \textbf{Time of tasks occurrence}: When tasks are being introduced in the system that is also an important parameter. It means how many tasks were present in the beginning and when and how the rest are introduced.
    
    We assume that the tasks may be available to the robots in the following three possible ways. 
\end{itemize}

\begin{itemize}
    \item \emph{Static:} When all the tasks information is available with the robots from the beginning.
    
    \item  \emph{Continuous arrivals:} When the tasks' arriving rate is somewhat close to each other, i.e. in each round some tasks are introduced and the number of tasks introduced in the current round is close to the number of tasks introduced in the previous rounds.
    
    \item  \emph{Spiking arrival:} When the rate of occurrence of tasks are close at first and then a sudden hike in the rate occurs for a few rounds and this keeps repeating all tasks are introduced. 
\end{itemize}

Without loss of generality, we assume that the execution time is equal to the Euclidean distance between the current location of the robot and the task location. The metrics used for the evaluation of the proposed strategies are summarized below.

\begin{itemize}[]
    \item \textbf{Makespan time}: This can be defined as the overall time taken for the execution of all the tasks by the multi-robot system. It depends on how well the tasks have been distributed among the robots. A good allocation is indicated by the lesser makespan. 
    \item \textbf{Convergence time}: It is defined as the total time taken for the allocation algorithm to converge to allocate all the tasks among the available robots.
	\item \textbf{Total radio-on time}: It is defined as the total time for which the radio in the robots was turned on for transmitting/receiving to/from other robots during the allocation process.
\end{itemize}

\subsection{Task pattern}
We evaluate the three strategies discussed in Section~\ref{sec:design} for a scenario when all the tasks are known beforehand to all the robots, i.e., static scenario. However, in practice, all tasks may not be available at the beginning. Thus, we also consider two dynamic scenarios -- \textit{continuous} and \textit{spiking} as suggested in~\cite{agarwal2019cannot}. In the continuous scenario, at every fixed interval, a similar number of tasks are discovered/added to the system. In the spiking scenario, there are only a few tasks (or no task) that are added after most of the intervals, and a relatively large number of tasks are introduced to the system at the end of some of the intervals. For our experiments, we define 2000 ms as the task discovery interval. As each Chaos round takes about 2000 ms to complete, this duration ensures that many/all tasks are not available to the system before a Chaos round and they are gradually considered for allocation. There are roughly 4 tasks added to the system at every interval for the continuous scenario. In the spiking scenario, there are sets of 5 intervals, where only 1 or 2 tasks are added after the first three intervals and then 8 tasks are added for the next two intervals.

\subsection{Evaluation results}
Experiments have been carried out to justify our claims. We have run the experiments for all the different task patterns.

\subsubsection{Fast and energy-efficient consensus}
\begin{figure*}
    \centering
     \begin{subfigure}[b]{0.6\textwidth}
         \centering
         \includegraphics[width=\textwidth]{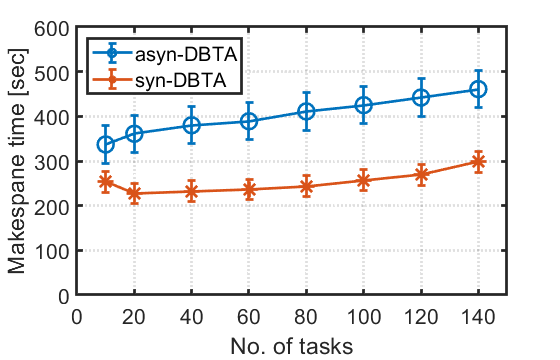}
         \caption{static arrival of tasks}
         \label{fig:stsyn}
     \end{subfigure}\\
     \begin{subfigure}[b]{0.6\textwidth}
         \centering
         \includegraphics[width=\textwidth]{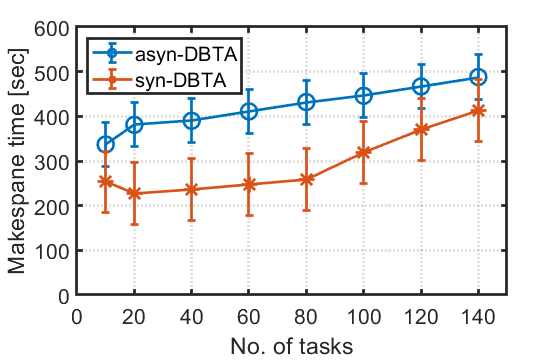}
         \caption{spiking arrival of tasks}
         \label{fig:spsyn}
     \end{subfigure}\\
     \begin{subfigure}[b]{0.6\textwidth}
         \centering
         \includegraphics[width=\textwidth]{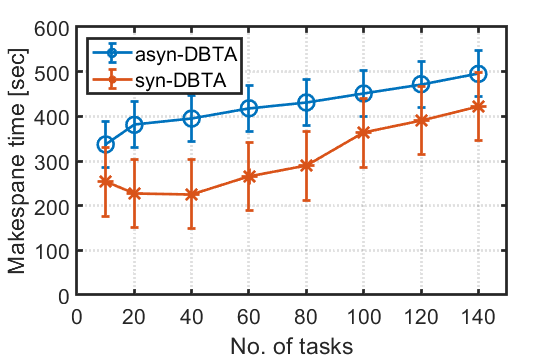}
         \caption{continuous arrival of tasks}
         \label{fig:cosyn}
     \end{subfigure}
    \caption{Comparison of makespan time between syn-DBTA (uses ST, i.e., Chaos) and asyn-DBTA (uses AT, i.e., netflood  \cite{netflood} mode of transfer while communication) for a) static, b) spiking and c) continuous arrival of tasks. All the experiments are carried out with 25 robots with the tasks distributed in an area of size 1000X1000 sq meters.}
    \label{fig:sync_three_graph}
\end{figure*}

First, we compare our algorithm DBTA which uses ST with an AT-based protocol - netflood \cite{netflood} to show the efficiency of using ST-based for communication. The netflood primitive sends a single packet to all nodes in the network. It broadcasts at every hop to reduce the number of redundant transmissions. It sets the end-to-end sender and end-to-end packet ID attributes on the packets it sends. A forwarding node saves the end-to-end sender and packet ID of the last packet it forwards and does not forward a packet if it has the same end-to-end sender and packet ID as the last packet. This reduces the risk of routing loops but does not eliminate them entirely as the netflood primitive saves the attributes of the latest packet seen only. Therefore, the netflood primitive also uses the time to live attribute, which is decreased by one before forwarding a packet. After each period the robots transfer the information regarding the bids using this protocol where each robot broadcasts at a single time. This increases the overall time to communicate and thus makespan time increases. Figure \ref{fig:sync_three_graph} shows the comparison of the makespan time for all the two versions of DBTA. 

\subsubsection{Fast convergence}
\begin{figure*}
    \centering
     \begin{subfigure}[b]{0.6\textwidth}
         \centering
         \includegraphics[width=\textwidth]{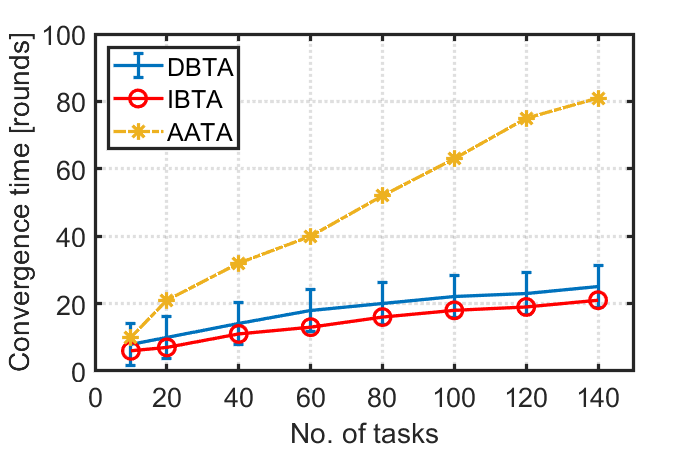}
         \caption{static arrival of tasks}
         \label{fig:stmk10_conv}
     \end{subfigure}\\
     \begin{subfigure}[b]{0.6\textwidth}
         \centering
         \includegraphics[width=\textwidth]{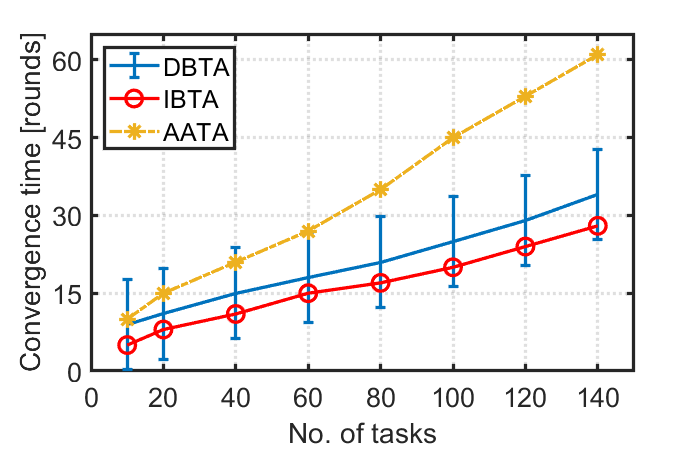}
         \caption{spiking arrival of tasks}
         \label{fig:stcn10_conv}
     \end{subfigure}\\
     \begin{subfigure}[b]{0.6\textwidth}
         \centering
         \includegraphics[width=\textwidth]{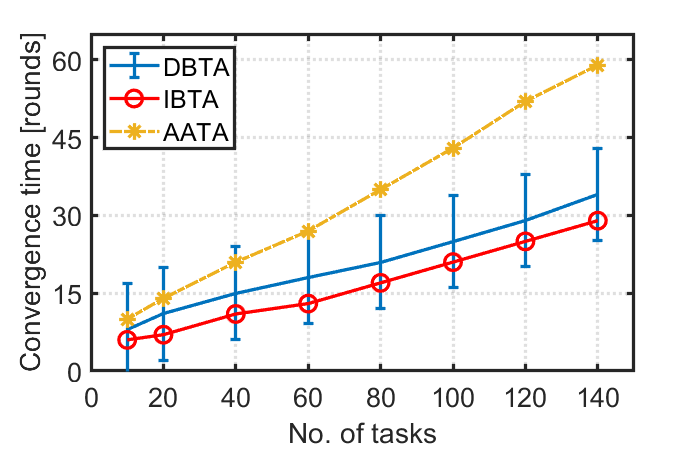}
         \caption{continuous arrival of tasks}
         \label{fig:stro10_conv}
     \end{subfigure}
    \caption{Comparison of convergence time among DBTA, IBTA, and AATA for a) static, b) spiking, and c) continuous arrival of tasks. All the experiments are carried out with 25 robots with the tasks distributed in an area of size 1000X1000 sq meters.}
    \label{fig:convergence_three_graph}
\end{figure*}

As mentioned earlier, we varied the number of tasks from 10 to 140, and for each setup, we run 15  iterations where task location varies across the iterations. Figure~\ref{fig:convergence_three_graph} shows the comparison of convergence time among the three proposed strategies for the static, continuous, and spiking scenarios. Under static task scenarios, in terms of convergence time, IBTA and DBTA perform similarly as they use fast in-network consensus-based strategies that need much less data exchange. However, AATA incurs a high convergence as well as radio-on time as it deals with all-to-all data sharing. Under the continuous task scenario, the convergence time or the number of rounds required by DBTA is quite similar to IBTA due to the approximately same quantity of information shared whereas for AATA the convergence time is too high due to the high amount of information exchange. The behavior of the spiking task scenario is quite similar to the set of experiments run with the tasks arriving continuously. The total convergence time for the allocation process of our algorithm is approximately similar to IBTA and is far better than AATA as it takes a large number of rounds to converge due to the high quantity of information shared.

\subsubsection{Minimal makespan time}
\begin{figure*}
    \centering
     \begin{subfigure}[b]{0.6\textwidth}
         \centering
         \includegraphics[width=\textwidth]{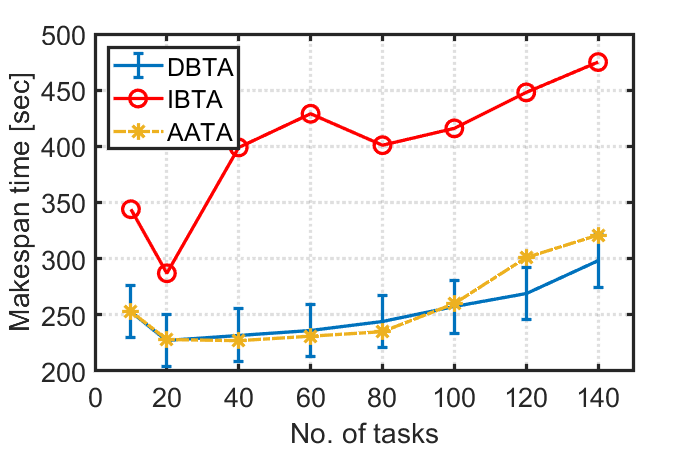}
         \caption{static arrival of tasks}
         \label{fig:stmk10_makespan}
     \end{subfigure}\\
     \begin{subfigure}[b]{0.6\textwidth}
         \centering
         \includegraphics[width=\textwidth]{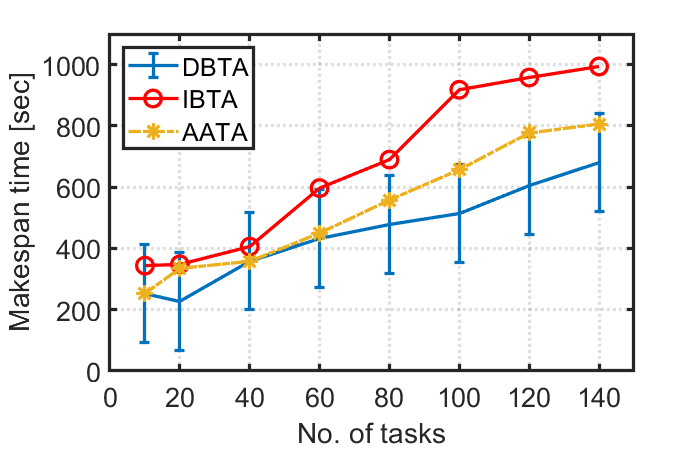}
         \caption{spiking arrival of tasks}
         \label{fig:stcn10_makespan}
     \end{subfigure}\\
     \begin{subfigure}[b]{0.6\textwidth}
         \centering
         \includegraphics[width=\textwidth]{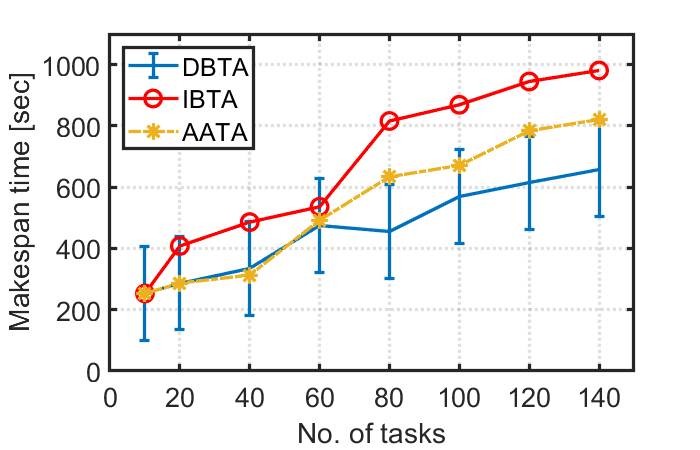}
         \caption{continuous arrival of tasks}
         \label{fig:stro10_makespan}
     \end{subfigure}
    \caption{Comparison of makespan time among DBTA, IBTA, and AATA for a) static, b) spiking, and c) continuous arrival of tasks. All the experiments are carried out with 25 robots with the tasks distributed in an area of size 1000X1000 sq meters.}
    \label{fig:makespan_three_graph}
\end{figure*}

Figure~\ref{fig:makespan_three_graph} shows the comparison of makespan time among the three proposed strategies for the static, continuous, and spiking scenarios. In the static scenario, DBTA and AATA achieve a similar makespan while IBTA shows a very high makespan due to poor allocation. While under the continuous task scenario, the makespan time for DBTA or the quality of solution shows a good improvement over IBTA and AATA due to better task allocation compared to the other two. Under the spiking task scenario, the makespan or the quality of solution for our algorithm DBTA shows good improvement over IBTA and AATA. 

\subsubsection{Less energy consumption}
\begin{figure*}
    \centering
     \begin{subfigure}[b]{0.6\textwidth}
         \centering
         \includegraphics[width=\textwidth]{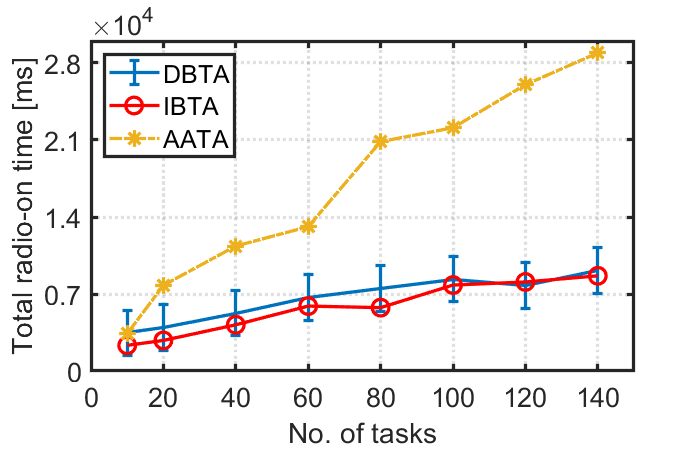}
         \caption{static arrival of tasks}
         \label{fig:stmk10}
     \end{subfigure}\\
     \begin{subfigure}[b]{0.6\textwidth}
         \centering
         \includegraphics[width=\textwidth]{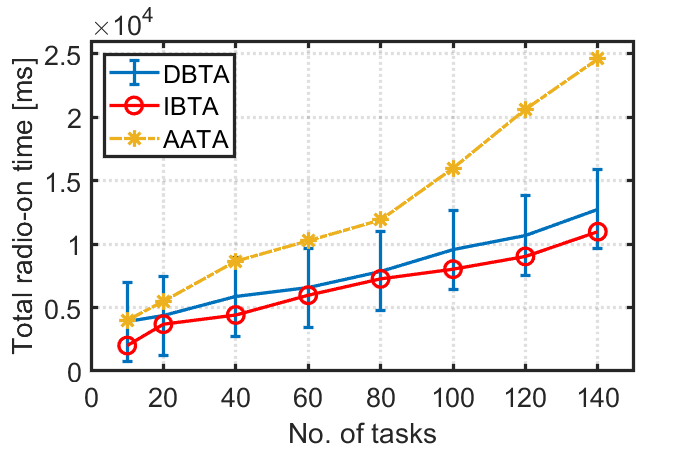}
         \caption{spiking arrival of tasks}
         \label{fig:stcn10}
     \end{subfigure}\\
     \begin{subfigure}[b]{0.6\textwidth}
         \centering
         \includegraphics[width=\textwidth]{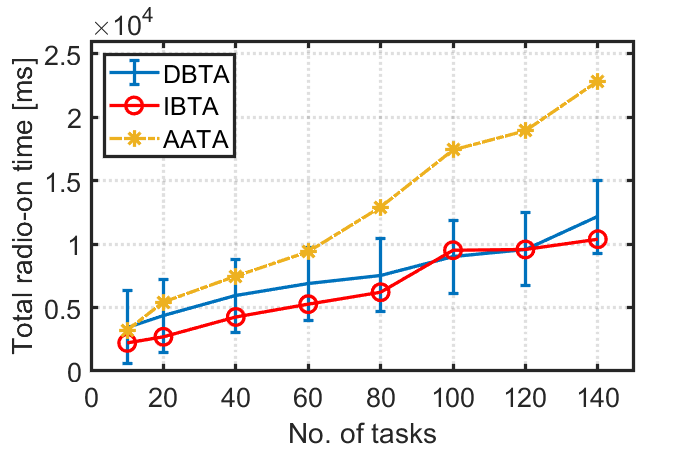}
         \caption{continuous arrival of tasks}
         \label{fig:stro10}
     \end{subfigure}
    \caption{Comparison of total radio-on time among DBTA, IBTA, and AATA for a) static, b) spiking, and c) continuous arrival of tasks. All the experiments are carried out with 25 robots with the tasks distributed in an area of size 1000X1000 sq meters.}
    \label{fig:energy_three_graph}
\end{figure*}

Figure~\ref{fig:energy_three_graph} shows the comparison of total radio-on time among all proposed strategies under all three different scenarios. In terms of radio-on time, IBTA and DBTA perform similarly as they use fast in-network consensus-based strategies that need much less data exchange. However, AATA incurs a high radio-on time because of higher convergence as it deals with all-to-all data sharing. Under the continuous task scenario, the number of rounds required by DBTA is quite similar to IBTA due to the approximately same quantity of information shared whereas for AATA the convergence time is too high due to the high amount of information exchange thereby reducing the total energy required, i.e., the total radio on time is approximately equal for both DBTA and IBTA whereas AATA requires significantly more amount of energy to converge. Under the spiking task scenario, the total energy required for the allocation process for our algorithm is approximately similar to IBTA and are far better than AATA as it takes a large number of rounds to converge, and consequently, a huge amount of energy is consumed due to the high quantity of information shared.

\subsubsection{Effect of more bidding}
\begin{figure*}
    \centering
     \begin{subfigure}[b]{0.6\textwidth}
         \centering
         \includegraphics[width=\textwidth]{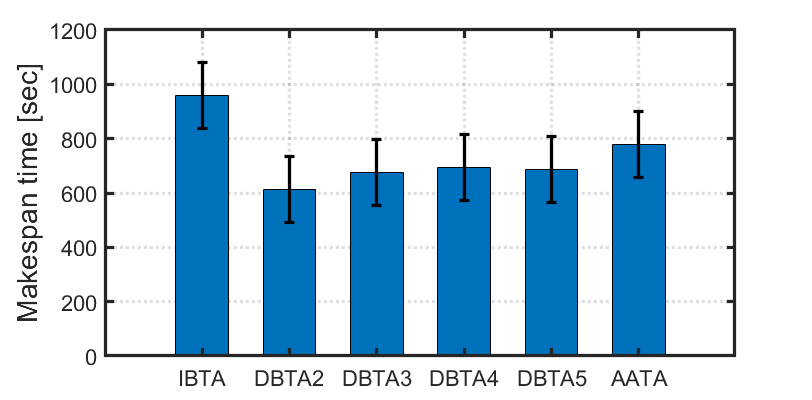}
         \label{fig:spmk25}
     \end{subfigure}\\
     
     \begin{subfigure}[b]{0.6\textwidth}
         \centering
         \includegraphics[width=\textwidth]{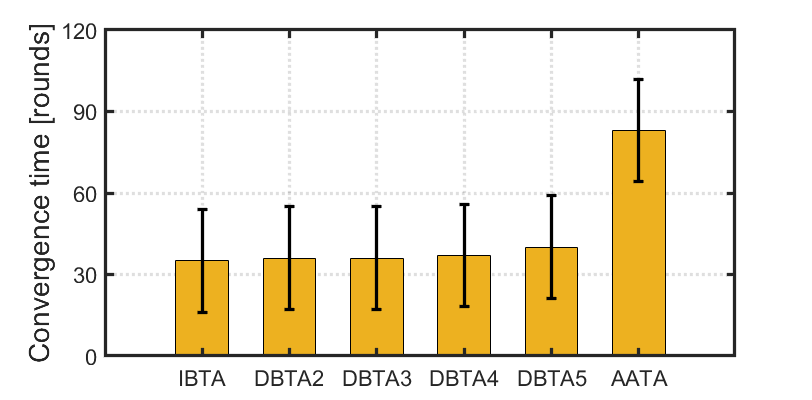}
         \label{fig:spcn25}
     \end{subfigure}
    \caption{Effect of multiple-bidding on makespan and convergence time for 140 tasks and 10 robots under continuous arrival of tasks.}
    \label{fig:multiple-bidding}
\end{figure*}

As mentioned in Section~\ref{sec:DBTA}, in the default setup DBTA uses 2 bid values per sharing. We run the experiment by increasing the number of bids shared to 3, 4, and 5 bids that are marked as DBTA3, DBTA4, and DBTA5, respectively. For the increased number of bid values, the algorithm is modified by extending the priority values to the number of bid values as described in Section~\ref{sec:algo}. These three schemes are compared with DBTA2, IBTA, and AATA, where DBTA2 is the same as DBTA. Figure~\ref{fig:multiple-bidding} showcases the result comparing the effect of multiple-biding on makespan and convergence time for 140 tasks and 10 robots. All the experiments are run for 15 iterations where the task location varies across iterations. The bar plot represents the average and standard deviation of these iterations. The results show that the quality of the solution (makespan) for DBTA2, DBTA3, DBTA4, and DBTA5 are fairly comparable and perform better than IBTA and AATA. As expected the convergence time increases as the number of bids increases. But it will be approximately similar for IBTA and DBTA as they share 1-bid and 2-bid respectively. Overall we can say that DBTA shows good results in terms of quality of solution and converge time with 2-bids.

IEEE 802.15.4 defines a maximum packet size of 128 bytes, of which 125 bytes can be used for actual data. Our Chaos implementation has an 8-byte header. That is, if the payload is, say, 10 bytes, it supports up to 856 nodes since at most [856/8]= 107 bytes are available for flags.
Our packet structure contains [sequence no-4 bytes, task id-2 bytes, winner-2 bytes, priority-2 bytes, payload-4 bytes (each bid value of 2B), flags-103 bytes]. Thus to maximize the number of nodes participating we use such payload length. As we start sharing more than 2-bids, more than 1 packet is generated and thus takes more time to converge thereby increasing the makespan time. When we change the packet size by increasing the number of bid values it can carry thereby decreasing the number of nodes that can participate, i.e. [sequence no-4 bytes, task id-20 bytes, winner-20 bytes, priority-20 bytes, payload-40 bytes (each bid value of 2B), flags-13 bytes], it can be seen that the red line goes straight indicating there is not much fluctuation in makespan time by sharing 2,3,4,5 bids respectively. Whereas the packet used in our study carrying a maximum of 2-bids at a time shows an increase in makespan time as the number of bid values increases because more packets are needed to flood to all nodes, which is indicated by the blue line. In any of the cases IBTA- single bid sharing, the makespan time will always be higher because of the bad allocation of tasks. Thus we can conclude that our algorithm shows the optimal results when the packet size is utilized at its best by sharing the maximum number of bids it can carry, which in our case is 2-bid allowing more nodes to participate.
\section{Discussions}\label{sec:discussion}
In this section, we provide some discussion points that would be helpful to adapt the proposed work and perform future research.

\textbf{Theoretical studies on ST}: ST-based strategies, unlike AT-based strategies, carry out the assigned jobs much faster. Communication under ST happens in a more systematic way. However, there have been very few theoretical studies on such strategies. The work by Wang \textit{et al.}~\cite{wang2018analysis} does one such analysis which can be used to understand the underlying large-scale dynamics of the protocols. The work concludes with certain guidelines for the improvement of the performance of the algorithms under denser network settings as well as limitations in the concurrent transmitters. A similar analysis is applicable for the strategy used in the current work too. However, specific analysis can be carried out which is beyond the scope of the current work and we consider the same as our possible future work in this direction.

\textbf{Centralized vs. Decentralized}: We use the simulator Cooja to carry out our experiments. All the algorithms used in this work are distributed in nature. There is no node assumed to be doing any central coordination. The used simulator (Cooja) provides two ways to control the nodes. First, through the GUI of the simulator, the nodes can be controlled or the java code of the simulator can be extended to have more central control over the nodes created. Second, the nodes can talk to each other through their virtualized communication interface. Thus the nodes can control their activities mutually through talking to each other. In our experiment, except for the initialization process, we do not exploit the first strategy for controlling the activities of the nodes. We completely depend on the inter-node interaction to govern their activities. Thus, we depend on fully decentralized coordination.

\textbf{Hardware emulation}: It is also important to note that the Cooja platform extensively uses hard emulators. We test our proposed strategy on the emulated version of the real devices only. Specifically, we have used the hardware device known as Sky motes for communication purposes. Note that the code developed for cooja is fully portable to real hardware devices. Hence, we test our proposed strategy in real TelosB motes (sky motes) in our lab. Moreover, we also develop a small number (5) of robocar kits powered by RPi where we add the TelosB devices through a USB port of the RPIs. We run the developed protocol in those lab-made tiny robots and found them to be working as expected. However, because of a small number of devices, we do not include the results in the main manuscript.

\textbf{ST/Wifi/ZigBee}: ST-based strategies are becoming very popular because of their ability to operate at high speed with high reliability. ST-based strategies were originally developed with ZigBee radios. However, later they have been successfully ported to various other radio devices such as BLE, LoRa, UWB, etc. There is also support for various different microcontroller architectures. Thus, diversity in the radio and microcontroller architecture is not going to affect the feasibility of the proposed strategy.

\textbf{Micro-second level time synchronization}: Tight time synchronization, preferably at the level of a microsecond, is one of the main needs for ST to work. Kindly note that microsecond level time synchronization has been shown to be possible by many of the existing works and hence such strategies were possible to be ported to a variety of different platforms seamlessly. Such tight time synchronization is achieved with the help of communication of radio packets among the participating nodes. Thus, whatever radio-technology or microcontroller architecture it is, the necessary level of time-synchronization does not become a bottleneck for the proposed strategy.

\section{Conclusion and Future Works}
\label{sec:conclusion}

Distributed task allocation is the only feasible option in many multi-robot applications and an efficient communication protocol plays a crucial role in such scenarios. It not only ensures faster and superlative task allocation, but also plays a role in the energy consumption of the robot due to frequent and large amounts of information sharing. Taking a cue from the superior ST-based communication protocols in IoT/networked embedded applications, we demonstrate an efficient utilization of ST for distributed MRTA. To understand how we can exploit ST for task allocation purposes, we design three strategies - IBTA, DBTA, and AATA that combine a bidding-based task allocation algorithm. The proposed strategies fundamentally differ in how much bidding information the robots would share with each other and subsequently exploit the information cleverly for task allocation. In particular, IBTA shares only a single \textit{best} bid while DBTA shares two such bids with all the other robots. AATA is a naive strategy where all the robots share all the bidding data with each other. We first demonstrate, how the performance of such decentralized task allocation strategies degrades if the inter-robot communication is done through an AT-based mechanism compared to ST. Through extensive simulation-based evaluation, we compare the quality of the task allocation, time an allocation process takes to converge as well as the radio-on time consumed in the robots for each of these strategies. Through various experiments over three different types of task arrival patterns, e.g., static, continuous, and spiking, we demonstrate that DBTA achieves the best quality of the task allocation while consuming similar convergence time and radio-on time as IBTA. 

Our work demonstrates a novel way of facilitating multiple robots to interact with each other for carrying out a distributed task allocation by using ST. There are many applications of multi-robot systems which can get immensely benefited by using the proposed strategy. However, real-life applications may also encompass various hostile situations. For instance, the robots may be highly mobile and as a result, groups of robots may get disconnected from the rest. An important issue here would be how can these robots rejoin the network and participate in the task allocation process. While the current work is the first step towards application of ST in MRTA, it does not deal with all such cases. To efficiently handle such situations a more robust and dynamic framework for all-to-all/many-to-many data-sharing would be an inevitable component. ST based technologies are still in evolving stage. Many new protocols are being designed and developed such as ByteCast \cite{bytecast}, MiniCast and its variations \cite{saha2017efficient,lcn_madhav, flexicast_dssrg}, which are capable of serving the goals in a better way. Especially, under a considerably large and possibly segregated multi-robot setting, allocation of tasks are necessary to be done more based on local decisions. Several recent application of the ST based protocols focus on this specific issue \cite{ev-dssrg,grid-dssrg-full,grid-dssrg,dssrg-pbft,dssrg-traffic-chandra,mpc_dssrg}. Clever lightweight separation of zones is also an important issue in such massive and hostile settings \cite{divcon_dssrg,sfd_dssrg,finegrain_dssrg,space-dssrg}. However, application of these available tools in the context of multi-robot task allocation would require special attention as we explored and demonstrated in this work. We aim to pursue these directions as an immediate future step.

\bibliographystyle{elsarticle-num-names}
\bibliography{multi-robot-comm}







\end{document}